\documentclass[conference]{IEEEtran}
\IEEEoverridecommandlockouts

\usepackage{cite}
\usepackage{amsmath,amssymb,amsfonts}
\usepackage{algorithmic}
\usepackage{graphicx}
\usepackage{textcomp}
\usepackage{xcolor}
\usepackage{booktabs,makecell, multirow, tabularx}
\def\BibTeX{{\rm B\kern-.05em{\sc i\kern-.025em b}\kern-.08em
    T\kern-.1667em\lower.7ex\hbox{E}\kern-.125emX}}
\begin{document}


\title{\fontsize{18}{18}\selectfont Point Cloud Resampling with Learnable Heat Diffusion}

\author{
    \IEEEauthorblockN{Wenqiang~Xu\IEEEauthorrefmark{1}, Wenrui~Dai\IEEEauthorrefmark{1}, Duoduo~Xue\IEEEauthorrefmark{2}, Ziyang~Zheng\IEEEauthorrefmark{2}, Chenglin~Li\IEEEauthorrefmark{2}, Junni~Zou\IEEEauthorrefmark{1}, Hongkai~Xiong\IEEEauthorrefmark{2}}
    \IEEEauthorblockA{
        \IEEEauthorrefmark{1}Department of Computer Science and Engineering, Shanghai Jiao Tong University, Shanghai, China}
    \IEEEauthorblockA{
        \IEEEauthorrefmark{2}Department of Electronic Engineering, Shanghai Jiao Tong University, Shanghai, China}
}



\maketitle

\begin{abstract}
Generative diffusion models have shown empirical successes in point cloud resampling, generating a denser and more uniform distribution of points from sparse or noisy 3D point clouds by progressively refining noise into structure. However, existing diffusion models employ manually predefined schemes, which often fail to recover the underlying point cloud structure due to the rigid and disruptive nature of the geometric degradation.
To address this issue, we propose a novel learnable heat diffusion framework for point cloud resampling, which directly parameterizes the marginal distribution for the forward process by learning the adaptive heat diffusion schedules and local filtering scales of the time-varying heat kernel, and consequently, generates an adaptive conditional prior for the reverse process. Unlike previous diffusion models with a fixed prior, the adaptive conditional prior selectively preserves geometric features of the point cloud by minimizing a refined variational lower bound, guiding the points to evolve towards the underlying surface during the reverse process.
Extensive experimental results demonstrate that the proposed point cloud resampling achieves state-of-the-art performance in representative reconstruction tasks including point cloud denoising and upsampling.
\end{abstract}

\begin{IEEEkeywords}
Point cloud resampling, diffusion model, learnable heat diffusion.
\end{IEEEkeywords}

\section{Introduction}
Point clouds are irregularly sampled from surfaces of 3D objects and usually suffer from noise and low density due to inherent limitations of scanning devices or ambiguities in reconstruction from images. Point cloud resampling aims to enhance the quality for reconstruction and understanding by reducing noise and increasing density. It is challenging but essential for 3D vision applications like autonomous driving and robotics. Reliable techniques are demanded for tackling the irregular nature of point clouds. 

Optimization-based methods have been extensively studied for solving point cloud resampling, including techniques such as grid-based resampling, moving least squares (MLS) surface theory, and graph-based resampling. These methods generally rely on geometric priors~\cite{b5,b6}, yet balancing the preservation of fine details with computational efficiency remains a significant challenge.
Grid-based resampling simplifies point clouds using structures like Poisson disk sampling~\cite{b1} and kd-trees~\cite{b2}. Its advantages include simplicity, low computational cost, and approximate uniform density control. However, it could lose geometric features such as sharp edges, fail in isotropic resampling without local detection, and lack geometric consistency. 
In addition to grid-based resampling, Moving Least Squares (MLS) surface theory~\cite{b3} provides an effective approach for maintaining geometric consistency during resampling. MLS defines a continuous surface based on local regions to ensure the resampled surface fits the original point cloud. However, MLS-based methods still face challenges with intrinsic control, causing errors in areas with sharp curvature. Graph-based resampling approaches~\cite{b4} offer another alternative that utilizes edges to model pairwise relationships between points and partially represent the multilateral interactions in 3D data. However, these methods often struggle to fully capture the complex geometries of point clouds due to limitations in bilateral connections. Efficient graph construction for arbitrary point clouds remains a significant challenge. 

In recent years, deep learning-based approaches have garnered significant attention in point cloud resampling, achieving remarkable performance by developing network architectures specifically designed for point clouds~\cite{b7}.
These methods usually learn to correct or adjust point positions and refine the distribution of points to better align with the underlying structure or pattern. However, they could suffer from degraded resampling quality due to artifacts like point shrinkage or excessive dispersion caused by over-correction~\cite{b26} or under-correction~\cite{b15}. Although this issue could be mitigated by iteratively adjusting point positions to align with the estimated surface~\cite{b8}, the process tends to become unstable under discontinuous gradients and insufficient regularization terms for optimization. Complex regularization terms~\cite{b9} and meticulous fine-tuning are often required to prevent trivial patterns such as clustered or overly uniform distributed points.

Lately, generative diffusion models~\cite{b10} have demonstrated their ability to generate high-quality data by progressively refining noise into structure, and have been further extended by incorporating various degradation techniques, such as blurring and masking, to expand their generative potential~\cite{b11,b12}. A heat diffusion-based method tailored specifically for 3D point clouds~\cite{b13} enables gradient detection within dense data distributions and refines the accuracy of interpolated points while preserving complex geometric details. However, these methods are constrained by manually predefined degradation schemes. 
For instance, in noise perturbation models, both the noise intensity schedule and the number of steps are predefined. Similarly, in heat diffusion models, the heat kernel, time integration interval, and discrete step size for solving the PDEs are manually set. These predefined schemes result in a fixed conditional prior that often fails to effectively guide the recovery of the underlying point cloud structure due to the rigid and disruptive nature of the geometric degradation.
Although recent diffusion models~\cite{b14, b27} have explored parameterizing the forward process, they focus on 2D tasks using simple noise perturbation strategies. In contrast, more flexible degradation approaches that could better address the complexities of 3D point clouds have yet to be fully explored.

In this paper, we propose a novel point cloud resampling method using learnable heat diffusion. Unlike existing approaches~\cite{b12, b13} that employ fixed heat diffusion in the forward process, we leverages learnable heat diffusion to directly parameterize the marginal distribution produced by the forward process, generating an adaptive conditional prior for the reverse process. Different from predefined and uniform degradation schemes, the adaptive conditional prior selectively preserves geometric features of the point cloud, guiding the points to evolve towards the underlying surface during the reverse process. Experimental results demonstrate the effectiveness of the proposed method, especially in handling low-density and noisy point clouds.

The contributions of this paper are summarized as below.
\begin{itemize}
\item We develop a novel diffusion model-based framework for point cloud resampling that uses learnable heat diffusion for the forward process and enables an adaptive conditional prior for the reverse process. The proposed framework facilitates the resampled point clouds to approach the underlying surface.
\item We propose a learnable heat diffusion without manually tuning hyperparameters for the forward diffusion process. We learn the integration time interval, discrete step size, and local filtering scale of the time-varying heat kernel to achieve adaptive conditional prior for preserving geometric features of irregular point clouds.
\item We design a refined variational lower bound (VLB) based on the adaptive conditional prior.
By minimizing the refined VLB, we enable accurate resampling using the approximation from conditional distribution based on the low-quality point clouds. 
\end{itemize}


\section{Proposed Method}\label{methodology}


In this section, we first present the motivation behind our point cloud resampling method. Subsequently, we elaborate on the learnable heat diffusion process and develop the corresponding refined variational lower. Finally, we cover the practical aspects of both training and inference.


\subsection{Motivation}\label{Motivation}
In this paper, we focus on estimating the conditional distribution \(p(\mathbf{x}_h | \mathbf{x}_l)\) of high-quality version~\(\mathbf{x}_h \in \mathbb{R}^{3}\) from low-quality input point clouds~\(\mathbf{x}_l \in \mathbb{R}^{3}\) for point cloud resampling. Learning-based methods leverage deep neural networks to improve \(p(\mathbf{x}_h | \mathbf{x}_l)\) but are challenged by non-linearities, data sparsity, irregular sampling, and noise in the low-quality data when directly modeling the complex distribution. Inspired by the success of generative diffusion models, we propose a novel diffusion model based framework for point cloud resampling that generates high-quality resampled point clouds by progressively transforming a simple distribution by corrupting the low-quality point cloud with learnable heat diffusion. 

Diffusion models degrade data to a fixed conditional prior \(\mathbf{z} \in \mathbb{R}^{3}\) (e.g., Gaussian noise) through a rigid, step-by-step process, leading to a significant degradation of the point cloud's geometric structure. This makes it challenging to accurately reconstruct geometric details during the reverse process.
In this paper, we accommodate to this concept by learning a degradation scheme tailored for point cloud resampling. Contrary to directly modeling \(p(\mathbf{x}_h | \mathbf{x}_l)\), we diffuse the low-quality point cloud towards a learnable conditional prior to approximate \(p(\mathbf{z} | \mathbf{x}_h)\) with \(p(\mathbf{z} | \mathbf{x}_l)\). 
We then solve the inverse process to recover the high-quality output.
Formally, this process is modeled through the joint distributions \(p(\mathbf{x}_l, \mathbf{z}) = p(\mathbf{z} | \mathbf{x}_l)p(\mathbf{x}_l)\) for the forward process and \(p(\mathbf{x}_h, \mathbf{z}) = p(\mathbf{x}_h | \mathbf{z})p(\mathbf{z})\) for the reverse process. The overall workflow for the complete resampling process is described by \(p(\mathbf{x}_h) = \int\int p(\mathbf{x}_h | \mathbf{z}) p(\mathbf{z} | \mathbf{x}_l) p(\mathbf{x}_l) \, d\mathbf{z}d\mathbf{x}_l\).

\subsection{Learnable Heat Diffusion}\label{Learnable Heat Diffusion}

We employ a heat diffusion process that operates on a point cloud \(\mathbf{X} = \left \{ \mathbf{x}_1,\cdots, \mathbf{x}_i, \cdots, \mathbf{x}_N\right\} \in \mathbb{R}^{N \times 3}\), where \(N\) represents the number of points. We define the latent variables \(\mathbf{z}^t\) as a sequence of progressively diffused versions of \(\mathbf{x}\) with \(t\) ranging from \(t = t_0\) (initial, least diffused) to \(t = t_T\) (final, most diffused). 
The evolution of \(\mathbf{z}^t\), conditioned on \(\mathbf{x}\) for \(t \in [t_0, t_T]\) is governed by the heat equation.
\begin{equation}
\frac{\partial \mathbf{z}(t)}{\partial t} = \nabla^2 \mathbf{z}(t),
\end{equation}
where \(\nabla^2 \in \mathbb{R}^{N \times N}\) is the Laplace operator for governing the diffusion process across the 3D surface. The conditional distribution for diffusion is represented as:
\begin{equation}\label{distribution trans}
p\left(\mathbf{z}^{t+\tau} | \mathbf{z}^{t}\right)= (\mathrm{I} - \tau \nabla^2 \mathbf{z}^{t})^{-1} \mathbf{z}^{t},
\end{equation}
The cumulative transformation over the time interval from \(t_0\) to \(t_T\) is formulated with a partial differential equation (PDE).
\begin{equation}\label{pde equation}
p(\mathbf{z}^{t_T} | \mathbf{x}) = p(\mathbf{z}^{t_0}| \mathbf{x}) + \int_{t_0}^{t_T} p\left(\mathbf{z}^{t+\tau } | \mathbf{z}^{t}\right) \, d\tau,
\end{equation}
where \(\tau\) represents the step size in the time-discretized form of the PDE.
In this framework, the Laplace operator \(\nabla^2\) is constructed in a time-varying manner using a graph representation of the point cloud, where each point is treated as a node in the graph. The adjacency matrix is built by finding the \(k\)-nearest neighbors for each node, and the edge weights are determined by a Gaussian radial basis function served as the heat kernel. Given a center point $i$, the heat kernel is defined as
\begin{equation}\label{Laplace operator}
\nabla^2(\mathbf{z}_i, t) = \sum_{j:(i, j) \in \xi} \frac{1}{\sqrt{2\pi \mathbf{\sigma}_{i}^2(t)}} \exp\left(-\frac{(\mathbf{z}_i - \mathbf{z}_j)^2}{2\mathbf{\sigma}_{i}^2(t)}\right),
\end{equation}
where \(j\) represents a neighboring point of point \(i\), \(\xi\) is the set of neighboring points, and \(\sigma_i(t)\) is a time-varying parameter that controls the scale of the Gaussian filter.

To this end, we propose to learn the step size \(\tau\), the time schedule \(t \in [t_0, t_T]\), and the scale of the Gaussian filter for solving \eqref{pde equation} as elaborated below.

\textbf{i) Step size \(\tau\).} 
When solving~\eqref{pde equation} with the differentiable \textit{Fourth-order Runge-Kutta} method~\cite{b16}, the performance and stability of the approximation are highly sensitive to the step size \(\tau\). Thus, we make \(\tau\) a learnable parameter to balance accuracy and computational cost.

\textbf{ii) Time schedule \(t \in [t_0, t_T]\).} We parameterize \(t = \mathbf{g}_{\eta}(t)\) using a a non-negative monotonic neural network~\cite{b14} \(\mathbf{g}_{\eta}(\cdot)\) with learnable parameters \(\eta\). The learned time schedule governs the overall intensity of the heat diffusion.

\textbf{iii) Scale of Gaussian filter.} We learn the scale \(\mathbf{f}(\mathbf{\sigma}_i(t))\) of the Gaussian filter for each patch using point-wise MLPs (multi-layer perceptrons) through a linear transformation. The learned scale \(\mathbf{f}(\mathbf{\sigma}_i(t))\) determines the extent to which geometric details are preserved in different regions of the point cloud, and controls the rate of heat diffusion across different regions and times in the point cloud. 

\subsection{Refined Variational Lower Bound}\label{Optimization of Variational Lower Bound}
Diffusion models are trained by optimizing the variational lower bound of \(-\log p(\mathbf{x})\), with the loss function given as:
\begin{equation}\label{vae}
\underbrace{D_{\mathrm{KL}}\left(q(\mathbf{z}^{t_{T}} | \mathbf{x}) \| p(\mathbf{z}^{t_{T}})\right)}_{\mathcal{L}_{t_T}}
\!+\! \underbrace{\left[-\log p(\mathbf{x} | \mathbf{z}^{t_0})\right]}_{\mathcal{L}_{t_0}}
\!+\! \underbrace{\mathcal{L}_{t}}_{\mathcal{L}_{t_0\sim t_T}(\mathbf{x})}.
\end{equation}
In \eqref{vae}, the prior loss \(\mathcal{L}_{t_T} \) and reconstruction loss \(\mathcal{L}_{t_0}\) are often underemphasized due to the fixed nature of the forward process. Most diffusion models focus on the following loss:
\begin{equation}
\mathcal{L}_{t}(\mathbf{x})= \sum_{t>t_0+\tau} D_{\mathrm{KL}}\left(q(\mathbf{z}^{t-\tau} | \mathbf{z}^{t}, \mathbf{x}) \| p_{\theta}(\mathbf{z}^{t-\tau} | \mathbf{z}^{t})\right).
\end{equation}
$\mathcal{L}_{t}(\mathbf{x})$ is challenging to optimize due to its dependence on the hyperparameter  \(t \in [t_0, t_T]\), which controls temporal granularity and generative depth.

The proposed learnable heat diffusion enables direct parameterization of \(q(\mathbf{z}^{T} | \mathbf{x})\). \(p(\mathbf{z} | \mathbf{x}_l)\) given the input \(\mathbf{x}_l\) can be transformed into \(p(\mathbf{z} | \mathbf{x}_h)\) given the corresponding ground truth \(\mathbf{x}_h\). The variational lower bound for optimization is then reformulated as
\begin{align}\label{Optimized vlb}
&D_{\mathrm{KL}}\left(q(\mathbf{z}^{t_{T}} | \mathbf{x}_h) \| p_{\theta}(\mathbf{z}^{t_{T}} | \mathbf{x}_l) \right) + \left[-\log p_{\theta}(\mathbf{x}_o | \mathbf{z}^{t_0}, \mathbf{x}_l)\right]  \nonumber\\
& \ + \sum_{t>t_0+\tau} D_{\mathrm{KL}}\left(q(\mathbf{z}^{t-\tau} | \mathbf{z}^{t}, \mathbf{x}_h) \| p_{\theta}(\mathbf{z}^{t-\tau} | \mathbf{z}^{t}, \mathbf{x}_l)\right).
\end{align}
Here, \(\mathbf{x}_o\) is the final output of the resampling process. 

The refined VLB formulation~\eqref{Optimized vlb} provides a more robust framework for optimization that integrates the reconstruction loss and prior loss with the diffusion process. It can dynamically adjust the distribution of latent variables to enable a smoother transition from low-quality inputs \(p(\mathbf{x}_l)\) to higher-quality outputs \(p(\mathbf{x}_h)\) and enhance resampling performance. 


\subsection{Training and Inference}\label{Training and Inference}


To enhance the stability of the training process for \eqref{Optimized vlb}, a small noise perturbation \(\delta\) is manually added to the conditional distributions \(q\left(\mathbf{z}^{t} |  \mathbf{x_h}\right)\) at each time step. This noise serves as an error tolerance or relaxation parameter~\cite{b12}, determining how similar two heat-diffused point clouds need to be to be considered indistinguishable. The modified conditional distributions are expressed as:
\begin{equation}
\tilde{q}\left(\mathbf{z}^{t} |  \mathbf{x_h}\right) =  \mathcal{N}\left(q\left(\mathbf{z}^{t} |  \mathbf{x_h}\right),\delta^{2}I\right).
\end{equation}
We train the model using widely used metrics for evaluating point cloud distribution, specifically a combination of Earth Mover's Distance (EMD) and Chamfer Distance (CD)~\cite{b15}, rather than relying on the KL divergence.

During the reverse inference process, the learned heat diffusion acts as a conditional prior \(p_{\theta}\left(\mathbf{z}^{t_{T}} | \mathbf{x}_l\right)\) at the initial time for low-quality input \(\mathbf{x}_l\), significantly improving the quality of point cloud resampling. The process is defined as:
\begin{align}
\mathbf{x}_o~\sim~&p_{\theta}\left(\mathbf{x}_{o} | \tilde{\mathbf{z}}^{t_{0}}, \mathbf{x}_{l}\right) \prod_{t} p_{\theta}\left(\mathbf{z}^{t-\tau} | \tilde{\mathbf{z}}^{t}, \mathbf{x}_{l}\right) p_{\theta}\left(\tilde{\mathbf{z}}^{t_{T}} | \mathbf{x}_{l}\right), \nonumber\\
&\text{with} \quad p(\tilde{\mathbf{z}} | \mathbf{x}_{l}) \sim \mathcal{N}\left(p(\mathbf{z}| \mathbf{x}_{l}), \delta^{2} \mathbf{I}\right).
\end{align}
The noise \(\delta\) controls stochastic sampling and defines the trajectory of point cloud resampling during inference.

\begin{table*}[!t]
\renewcommand{\baselinestretch}{1.0}
\renewcommand{\arraystretch}{1.0}
\setlength{\abovecaptionskip}{0pt}
\setlength{\belowcaptionskip}{0pt}

\caption{Results for \(4\times\) and \(16\times\) upsampling. Metrics are CD (\(\times 10^{-6}\)), HD (\(\times 10^{-4}\)), and EMD (\(\times 10^{-6}\)).}\label{results:table1}
\centering
\begin{tabular}{l|cccccc|cccccc}
\hline
                 & \multicolumn{6}{c|}{PU1K}                                                                                               & \multicolumn{6}{c}{PUGAN}                                                                                               \\ \hline
Up ratio & \multicolumn{3}{c}{4\(\times\)}                            & \multicolumn{3}{c|}{16\(\times\)}                          & \multicolumn{3}{c}{4\(\times\)}                            & \multicolumn{3}{c}{16\(\times\)}                           \\ \hline
Model            & CD\(\downarrow \) & HD\(\downarrow \) & EMD\(\downarrow \) & CD\(\downarrow \) & HD\(\downarrow \) & EMD\(\downarrow \) & CD\(\downarrow \) & HD\(\downarrow \) & EMD\(\downarrow \) & CD\(\downarrow \) & HD\(\downarrow \) & EMD\(\downarrow \) \\ \hline
PU-GAN           & 7.894             & 0.5788            & 13.57              & 5.261             & 0.7407            & 10.58              & 32.79             & 4.074             & 56.38              & 21.75             & 3.793             & 42.96              \\
PU-GCN           & 6.661             & 0.5054            & 19.10              & 6.048             & 0.6019            & 12.26              & 29.92             & 1.967             & 4.615              & 27.01             & 2.450             & 55.04              \\
Meta-PU          & 6.237             & \textbf{0.4199}   & 10.33              & 3.349             & \textbf{0.4172}   & 8.008              & 25.91             & 1.346             & 4.607              & 14.28             & 1.917             & 33.56              \\
Grad-PU          & 13.48             & 1.486             & 21.92              & 8,922             & 1.276             & 23.57              & 50.49             & 5.703             & 12.68              & 34.96             & 4.468             & 89.95              \\
NePs             & 5.195             & 0.445             & 9.232              & 3.287             & 0.563             & 7.249              & 2.332             & \textbf{1.304}    & 4.060              & 13.81             & \textbf{1.645}    & 29.29              \\ \hline
Ours         & \textbf{3.884}    & 0.432             & \textbf{7.199}     & \textbf{2.479}    & 0.528             & \textbf{6.760}     & \textbf{2.127}    & 1.310             & \textbf{3.353}     & \textbf{13.63}    & 1.693             & \textbf{26.65}     \\ \hline
\end{tabular}%
\end{table*}

\begin{table*}[!t]
\renewcommand{\baselinestretch}{1.0}
\renewcommand{\arraystretch}{1.0}
\setlength{\abovecaptionskip}{0pt}
\setlength{\belowcaptionskip}{0pt}
\centering
\caption{Results for denoising with noise levels of \(0.01\), \(0.02\), and \(0.03\). Metrics are CD (\(\times 10^{-6}\))and EMD (\(\times 10^{-6}\)).}\label{results:table2}
\begin{tabular}{l|cccccc|cccccc}
\hline
            & \multicolumn{6}{c|}{ModelNet~(\emph{Sparse})}                                                                            & \multicolumn{6}{c}{ModelNet~(\emph{Dense})}                                                                              \\ \hline
Noise level & \multicolumn{2}{c}{0.01}               & \multicolumn{2}{c}{0.02}               & \multicolumn{2}{c|}{0.03}              & \multicolumn{2}{c}{0.01}               & \multicolumn{2}{c}{0.02}               & \multicolumn{2}{c}{0.03}               \\ \hline
Model       & CD\(\downarrow \) & EMD\(\downarrow \) & CD\(\downarrow \) & EMD\(\downarrow \) & CD\(\downarrow \) & EMD\(\downarrow \) & CD\(\downarrow \) & EMD\(\downarrow \) & CD\(\downarrow \) & EMD\(\downarrow \) & CD\(\downarrow \) & EMD\(\downarrow \) \\ \hline
DGCNN       & 7.70              & 1.95               & 17.86             & 4.27               & 19.32             & 4.88               & 3.19              & 2.52               & 4.85              & 4.02               & 8.09              & 6.72               \\
DMR         & 11.19             & 2.66               & 13.70             & 3.34               & 17.43             & 4.65               & 4.39              & 3.65               & 5.77              & 4.51               & 8.25              & 6.80               \\
ScoreNet    & 6.71              & 2.13               & \textbf{9.53}              & 2.85               & 16.63             & 4.39               & 2.64              & 2.15               & 5.19              & 4.25               & 11.30             & 7.92               \\
Deeprs      & 6.02              & \textbf{1.21}               & 10.87             & \textbf{2.21}               & 15.78             & 3.31               & 3.70              & 2.37               & 7.26              & 4.74               & 14.19             & 8.76               \\
PathNet     & 9.99              & 2.16               & 12.98             & 2.88               & 19.11             & 4.26               & 8.14              & 5.49               & 9.66              & 7.17               & 14.70             & 10.75              \\ \hline
Ours    & \textbf{5.87}     & 1.87      & 10.37     & 2.41      & \textbf{15.36}             & \textbf{3.15}               & \textbf{2.13}     & \textbf{1.57}      & \textbf{4.15}     & \textbf{3.34}      & \textbf{6.95}     & \textbf{5.67}      \\ \hline
\end{tabular}%
\end{table*}

\begin{figure*}[!t]
    \centering
    \includegraphics[width=1\linewidth]{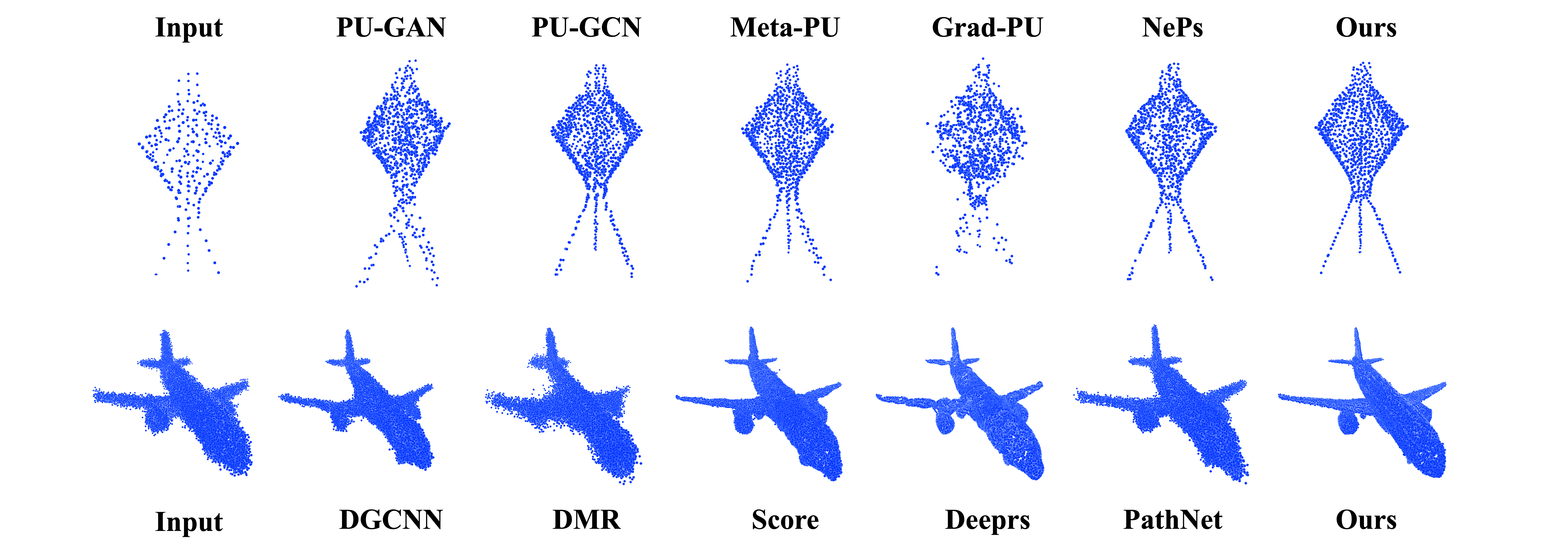}
    \caption{Comparison of resampling results across competitive benchmarks. The top row shows the 4x upsampling results for sparse point clouds, while the bottom row presents denoising results for noisy point clouds with a noise level of 0.01.}
    \label{fig:visual}
\end{figure*}

\section{Experimental Results}\label{experiments}
\subsection{Implementation Details}
\subsubsection{Dataset}  
We utilize three publicly available benchmark datasets—PU1K~\cite{b17}, PUGAN~\cite{b18}, and ModelNet~\cite{b20}—for both training and evaluation. We evaluate the performance of the proposed resampling model by applying it to two types of low-quality point cloud inputs: sparse and noisy point clouds. The training datasets, drawn from these three sources, collectively comprise 69,000 uniformly distributed patches. For testing, we extract 127 sparse point cloud samples for upsampling (evaluated at 4x and 16x upsampling levels) and 100 noisy point cloud samples for denoising, with Gaussian noise levels of 0.01, 0.02, and 0.03. These 100 noisy point cloud samples include both sparse and dense versions.

\subsubsection{Network}  
In addition to the learnable parameters introduced in Section~\ref{Learnable Heat Diffusion}, the overall framework includes a primary network responsible for learning the transformation of conditional distributions. For this purpose, we employ the point transformer~\cite{b19}, a backbone proven to be highly efficient in point cloud tasks. We further apply adaptive fine-tuning to tailor it to the specific requirements of our model.

\subsubsection{Baselines and Metrics}
We trained PU-GAN~\cite{b18}, PU-GCN~\cite{b17}, Meta-PU~\cite{b23}, Grad-PU~\cite{b24} and NePs~\cite{b25} as baselines for the upsampling task, and DGCNN~\cite{b26}, DMR~\cite{b15}, ScoreNet~\cite{b8}, Deeprs~\cite{b29}, and PathNet~\cite{b30} for comparison in the denoising task. Following prior works, we adopt Chamfer Distance (CD), Hausdorff Distance (HD), and Earth Mover’s Distance (EMD) as the evaluation metrics. For all metrics, smaller values indicate better quality of the results.

\subsubsection{Settings}  
The proposed model is trained over 300 epochs with a batch size of 16, targeting both upsampling and denoising tasks. We use the AdamW optimizer with an initial learning rate of 0.0002, and the noise parameter \(\delta\) is set to 0.01. All training and evaluation are performed using PyTorch on an NVIDIA GTX 4090 GPU.

\subsection{Comparison Results}

We unified the test results from PU1K~\cite{b17} and PUGAN~\cite{b18} to offer more comprehensive insights into upsampling performance, with quantitative results for \(4 \times\) and \(16 \times\) upsampling shown in Table~\ref{results:table1} and corresponding qualitative results in the top row of Figure~\ref{fig:visual}. Similarly, we merged the sparse and dense versions of the ModelNet~\cite{b20} results for denoising evaluation, with quantitative results for denoising at Gaussian noise levels of \(0.01\), \(0.02\), and \(0.03\) presented in Table~\ref{results:table2}, and qualitative results in the bottom row of Figure~\ref{fig:visual}.

\subsubsection{Quantitative Results}In Table~\ref{results:table1}, our method achieves the best overall performance for point cloud upsampling at both \(4\times\) and \(16\times\) rates on the PU1K~\cite{b17} and PUGAN~\cite{b18} datasets, although it slightly underperforms in HD compared to Meta-PU~\cite{b23} on PU1K and NePs~\cite{b25} on PUGAN. For denoising, as shown in Table~\ref{results:table2}, our method excels in the dense version of ModelNet~\cite{b20}, but underperforms in EMD compared to Deeprs~\cite{b29} at noise levels \(0.01\) and \(0.02\), and in CD compared to ScoreNet~\cite{b8} at the \(0.01\) noise level.

\subsubsection{Qualitative Results}  
The top row of Figure~\ref{fig:visual} shows the upsampling results for the fan-like object from PU1K~\cite{b17}, where our method achieves more accurate shape, particularly in finer details like the ``legs''. The bottom row presents the denoising results for the airplane object, highlighting that while DGCNN~\cite{b26}, DMR~\cite{b15}, and PathNet~\cite{b30} leave some outliers, and ScoreNet~\cite{b8} and Deeprs~\cite{b29} remove outliers at the cost of shape distortion, our approach effectively removes noise while retaining the original geometric details.

\section{Conclusions}
We propose a novel point cloud resampling method using learnable heat diffusion, introducing learnability to the heat equation's temporal parameters and spatial smoothing scale for precise diffusion control. The adaptive conditional prior, derived from the parameterized marginal distribution, enables accurate geometric modeling and optimizes the variational lower bound. Extensive experiments demonstrate state-of-the-art performance of the proposed method.

\vspace{12pt}

\end{document}